\documentclass{article}

\usepackage{microtype}
\usepackage{graphicx}

\usepackage{booktabs}

\usepackage{hyperref}

\usepackage[accepted]{icml2024}

\usepackage{amsmath}
\usepackage{amssymb}
\usepackage{mathtools}
\usepackage{amsthm}

\usepackage[capitalize,noabbrev]{cleveref}

\theoremstyle{plain}

\theoremstyle{definition}

\theoremstyle{remark}

\usepackage[textsize=tiny]{todonotes}

\usepackage{amsmath,amsfonts,bm}

\def\eqref#1{equation~\ref{#1}}
\def\Eqref#1{Equation~(\ref{#1})}

\def\1{\bm{1}}

\def\vb{{\bm{b}}}

\def\vx{{\bm{x}}}

\def\mW{{\bm{W}}}

\DeclareMathAlphabet{\mathsfit}{\encodingdefault}{\sfdefault}{m}{sl}
\SetMathAlphabet{\mathsfit}{bold}{\encodingdefault}{\sfdefault}{bx}{n}

\DeclareMathOperator*{\argmax}{arg\,max}

\usepackage[acronym, nohypertypes={acronym}]{glossaries}
\usepackage{subcaption}
\usepackage{multirow}
\usepackage{array} 
\usepackage{colortbl}
\usepackage{xspace}
\usepackage{rotating}
\usepackage{balance}
\newacronym{DNN}{DNN}{Deep Neural Network}
\newacronym{VNN}{VNN}{Verification-Friendly Neural Network}
\newacronym{MBP}{MBP}{Magnitude-Based Pruning}
\newacronym{SOP}{SOP}{Sparse Optimization Pruning}
\newacronym{FNN}{FNN}{Fully-connected Neural Network}
\newacronym{CNN}{CNN}{Convolutional Neural Network}
\newacronym{PGD}{PGD}{Projected Gradient Descent}

\icmltitlerunning{VNN: Verification-Friendly Neural Networks
           with Hard Robustness Guarantees}

\begin{document}

\twocolumn[
\icmltitle{VNN: Verification-Friendly Neural Networks
           with Hard Robustness Guarantees}

\icmlsetsymbol{equal}{*}

\begin{icmlauthorlist}
\icmlauthor{Anahita Baninajjar}{Lund}
\icmlauthor{Ahmed Rezine}{Linkoping}
\icmlauthor{Amir Aminifar}{Lund}

\end{icmlauthorlist}

\icmlaffiliation{Lund}{Department of Electrical and Information Technology, Lund University, Lund, Sweden,}

\icmlaffiliation{Linkoping}{Department of Computer and Information Science, Linköping University, Linköping, Sweden}

\icmlcorrespondingauthor{Anahita Baninajjar}{anahita.baninajjar@eit.lth.se}

\icmlkeywords{Robustness, Verifiability, Verification-Friendly Neural Networks, Formal Guarantees, Verification Frameworks, Deep Neural Networks, Post-training, Constrained Sparsification}

\vskip 0.3in
]

\printAffiliationsAndNotice{}

\begin{abstract}
Machine learning techniques often lack formal correctness guarantees, evidenced by the widespread adversarial examples that plague most deep-learning applications. This lack of formal guarantees resulted in several research efforts that aim at verifying \glspl{DNN}, with a particular focus on safety-critical applications. However, formal verification techniques still face major scalability and precision challenges. The over-approximation introduced during the formal verification process to tackle the scalability challenge often results in inconclusive analysis. To address this challenge, we propose a novel framework to generate \glspl{VNN}. We present a post-training optimization framework to achieve a balance between preserving prediction performance and verification-friendliness. Our proposed framework results in \glspl{VNN} that are comparable to the original \glspl{DNN} in terms of prediction performance, while amenable to formal verification techniques. This essentially enables us to establish robustness for more \glspl{VNN} than their \gls{DNN} counterparts, in a time-efficient manner.
\end{abstract}

\section{Introduction}
The state-of-the-art machine learning techniques suffer from lack of formal correctness guarantees. It has been demonstrated by a wide range of ``adversarial examples'' in the deep learning domain \cite{moosavi2016deepfool, kurakin2018adversarial, liang2022adversarial} and presents a major challenge in the context of safety-critical applications \cite{huang2018survey}. To address this challenge, several verification frameworks for \glspl{DNN} have been proposed in the state of the art \cite{Reluplex, MILP, Cav17, DeepPoly}. However, the majority of state-of-the-art verification techniques suffer from poor scalability. As a result, there have been several excellent initiatives to improve the scalability of the existing verification frameworks for \glspl{DNN} \cite{Marabou, singh2019beyond, safedeep}. 

In contrast to the majority of the state-of-the-art verification techniques, here, we propose a new generation of \glspl{DNN} that are amenable to verification, or simply verification-friendly. \acrfullpl{VNN} allow more efficiency, in terms of time, and more verified samples, while maintaining on-par prediction performance with their \gls{DNN} counterparts. The key observation is that most verification frameworks are based on over-approximation and this over-approximation accumulates/amplifies along the forward pass. To reduce the over-approximation, we define an optimization problem to obtain \glspl{VNN} by enforcing sparsity on the \glspl{DNN}, constrained to satisfy/guarantee performance/robustness requirements. As such, our framework allows verification tools to establish robustness for a wider range of samples. While the main goal of \glspl{VNN} is to be verification-friendly, our proposed framework also plays a regularization role, leading to more robust \glspl{VNN} owing to their sparsity. 

We evaluate our proposed framework based on the classical MNIST image dataset~\cite{mnist} and demonstrate that \glspl{VNN} are indeed amenable to verification, while maintaining accuracy as their \gls{DNN} counterparts. The verification time of \glspl{VNN} is always less than the original \glspl{DNN} (up to one-third in some cases) due to the sparsity of the \glspl{VNN}. At the same time, and more importantly, we observe that \glspl{VNN} allow verification of up to 76 times more samples, besides their advantage in terms of verification time efficiency.

To evaluate our framework on real-world safety-critical applications, we consider two datasets from medical domain, namely, CHB-MIT~\cite{shoeb_chb-mit_2010}, and MIT-BIH~\cite{mit-bih}, for real-time epileptic seizure detection and real-time cardiac arrhythmia detection, respectively. Failure to detect an epileptic seizure or a cardiac arrhythmia episode in time may have irreversible consequences and potentially lead to death \cite{panelli2020sudep,stroobandt2019failure}. For such safety-critical applications, we demonstrate that the number of verified samples with \glspl{VNN} is up to 24 and 34 times more than \glspl{DNN}, considering the CHB-MIT and MIT-BIH datasets, respectively.

Finally, we compare the performance of \glspl{VNN} with two state-of-the-art techniques, namely, \gls{MBP} and \gls{SOP}~\cite{manngaard2018structural}. \gls{MBP} and \gls{SOP} approaches aim at introducing sparsity, without taking the robustness requirements into consideration. As a result, the sparse network is not guaranteed to satisfy the robustness requirements. 
 The experiments confirm this explanation and show that \glspl{VNN} are up to 46, 19, and 27 times more verification-friendly than \gls{MBP} models on \glspl{DNN} trained on MNIST, CHB-MIT, and MIT-BIH datasets, respectively. Moreover, \glspl{VNN} are up to 51 times more verification-friendly than \gls{SOP} models trained on the MNIST dataset.

\section{\acrlongpl{VNN}}

In this section, we discuss our proposed framework to generate \glspl{VNN}. The framework comes with an optimization problem that takes a trained model as input (i.e., the original \gls{DNN}) and results in a \gls{VNN}. In the following, we introduce the optimization problem and explain the formulation of the objective function and constraints.

\subsection{\acrlongpl{DNN}}
Let us first formalize our notation for \glspl{DNN}. A \gls{DNN} is a series of linear transformations and nonlinear activation functions to generate outputs using trained weights and biases. We consider ${f}_{k}(.):R^{n_{k-1}}\to R^{n_{k}}$ as the function that generates values of the $k$th layer from its previous layer, where $n_{k}$ shows the number of neurons in the $k$th layer. Therefore, the values of the $k$th layer, ${\vx}^{(k)}$, are given by ${\vx}^{(k)} = {f}_{k}({\vx}^{(k-1)})=act({{\mW}^{(k)}} {\vx}^{(k-1)} + {{\vb}^{(k)}})$, such that ${\mW}^{(k)}$ and ${\vb}^{(k)}$ are the weights and biases of the $k$th layer and $act$ is the activation function. 

\subsection{Layer-Wise Optimization Problem}
Our optimization problem to obtain a \glspl{VNN} is performed layer-by-layer. Therefore, here, we focus on the $l$th layer of an $N$-layer \gls{DNN}. 
To obtain a \gls{VNN}, our aim is to minimize the number of non-zero elements of weight ${\tilde{\mW}^{(l)}}$ and bias ${\tilde{\vb}^{(l)}}$ of the $l$th layer of an $N$-layer \gls{DNN} to have a model that is more sparse while still accurate. To minimize the number of non-zero elements of ${\tilde{\mW}^{(l)}}$ and ${\tilde{\vb}^{(l)}}$, we consider the $L_0$ norm ${\lVert . \rVert}_{0}$, also known as the sparse norm, as the objective function. The $L_0$ norm counts the number of non-zero elements in a vector or matrix. Therefore, the objective function is ${\lVert \tilde{\mW}^{(l)} \rVert}_{0,0} + {\lVert \tilde{\vb}^{(l)} \rVert}_{0}$, to find the minimum number of non-zero elements in $\tilde{\mW}^{(l)}$ and $\tilde{\vb}^{(l)}$. In the following, we outline our proposed constrained optimization problem to obtain a \gls{VNN}: 
\begin{align}
&\min_{\tilde{\mW}^{(l)},\tilde{\vb}^{(l)}} {\lVert \tilde{\mW}^{(l)} \rVert}_{0,0} + {\lVert \tilde{\vb}^{(l)} \rVert}_{0}, \label{obj}\\
& \hspace{1.1em}  \textrm{s.t.}  \hspace{0.3cm}
\vx^{(k)} = f_k(\vx^{(k-1)}), \; \forall k \in \{1, \dots, N\} \label{lbl1}\\ 
& \hspace{1.1cm} \tilde{\vx}^{(l)} = \tilde{f}_l(\vx^{(l-1)}),  \label{lbl2}\\
& \hspace{1.1cm} \tilde{\vx}^{(m)} = f_m(\tilde{\vx}^{(m-1)}), \forall m \in \{l+1, \dots, N\},\label{lbl3}\\
& \hspace{1.1cm} {\lVert\tilde{\vx}^{(l)} - \vx^{(l)}\rVert}_{\infty} \leq \mathbf{{\epsilon}}, \label{lbl4}\\
& \hspace{1.1cm}  \argmax \vx^{(N)} = \argmax \tilde{\vx}^{(N)} = y, \;\label{lbl5} \\
& \hspace{1.15cm}  \forall \vx^{(0)}\in \bm{D}_{val}. \label{lbl6}
\end{align} \label{opt}

Consider the constraints presented in Equations (\ref{lbl1})--(\ref{lbl6}). \Eqref{lbl1} establishes the neurons' values of the original \gls{DNN} with the trained weights and biases. For setting up the neurons' values after optimization, we define $\tilde{f}_{l}(.): R^{n_{l-1}}\to R^{n_{l}}$ as a new nonlinear function that transfers neurons' values of the $(l-1)$th layer to the target layer $l$ using the optimized weights and biases. Therefore, $\tilde{\vx}^{(l)}$ is given by $\tilde{\vx}^{(l)} = \tilde{f}_{l}({\vx}^{(l-1)})=act({\tilde{\mW}^{(l)}} {\vx}^{(l-1)} + {\tilde{\vb}^{(l)}})$ in~\Eqref{lbl2}. The change of neurons' values in layer $l$ may result in different values for the neurons of the next layers, hence we introduce~\Eqref{lbl3}. ~\Eqref{lbl3} computes updated values for neurons in layers from $l+1$ to the end of the \gls{DNN}, which is layer $N$, using the new values of neurons in layer $l$. \Eqref{lbl4} ensures that $\tilde{\vx}^{(l)}$ takes a value within the neighborhood of $\epsilon$ around ${\vx}^{(l)}$ to avoid extreme changes and an excessive loss of accuracy. 
If $\epsilon = 0$, the value of each neuron is not allowed to change after optimization, i.e.,  $\tilde{\vx}^{(l)} = \vx^{(l)}$. If $\epsilon > 0$, then $\tilde{\vx}^{(l)}$ can be different from $\vx^{(l)}$ and the optimization has more freedom to find a more sparse \gls{VNN}. 

In~\Eqref{lbl5}, the optimization problem is constrained to keep the same class as the one derived by the \gls{DNN} prior to the optimization. Consider that the input ${\vx}^{(0)}$ belongs to class $c$. It means ${\vx}^{(N)}_{c}>{\vx}^{(N)}_{i}$ for all neurons $i \neq c$ in the last layer $l=N$. To obtain the same class after optimization, the same holds for $\tilde{\vx}^{(N)}_{i}$, i.e., $\tilde{\vx}^{(N)}_{c}>\tilde{\vx}^{(N)}_{i}$ for all neurons $i \neq c$, which is formulated in the context of our linear optimization problem. Moreover, introducing a constant robustness margin $M$ to this inequality converts it to $\tilde{\vx}^{(N)}_{c}>\tilde{\vx}^{(N)}_{i}+ M$ to provide hard guarantees for enhancing the robustness of the \gls{DNN}. However, for the simplicity of the presentation, we adopt the compact form of this constraint in \Eqref{lbl5} in the remainder of this paper. In~\Eqref{lbl5}, $y$ is the label of each $\vx^{(0)}$ in the validation set $\bm{D}_{val}$. All constraints need to hold for all inputs in the validation set $\bm{D}_{val}$ (\Eqref{lbl6}). In the optimization problem, we consider the validation set instead of the training set to avoid over-fitting in the \gls{VNN}.

\subsection{Relaxation of Objective Function}

The $L_0$ norm is expressed in the form of an $L_p$ norm, but to be precise, it is not a norm~\cite{l0norm}. However, optimizing the $L_0$ norm poses significant challenges due to the non-convex nature of the $L_0$ norm. Here, we elaborate on how to overcome this difficulty.

The $L_1$ norm also referred to as the Manhattan norm, is the tightest convex relaxation of the $L_0$ norm in convex optimization~\cite{zass2006nonnegative}. While the $L_0$ norm counts the number of non-zero elements of a matrix, the $L_1$ norm sums the absolute values of the elements. Being a convex function, the $L_1$ norm can be used in convex optimization problems, which is not possible with the $L_0$ norm. Therefore, the optimization objective is reformulated as follows to be convex:
\begin{equation}
\min_{\tilde{\mW}^{(l)},\tilde{\vb}^{(l)}} \quad  {\lVert \tilde{\mW}^{(l)} \rVert}_{1,1} + {\lVert \tilde{\vb}^{(l)} \rVert}_{1}.
\label{object}
\end{equation}

\subsection{Reformulation of Constraints}

The constraints of the optimization problem are nonlinear, which is caused by the activation functions in $\tilde{f}_l(.)$ and $f_m(.)$. Henceforth, we perform the analysis on the ReLU activation function, which is the most popular activation function in \glspl{DNN}, but our framework can be extended to any nonlinear activation function that can be presented in a piece-wise linear form.

Performing function $ReLU(x) = \max(0, x)$ results in two possible states for each neuron $x_i^{(l)}$. Suppose $f_{l,i}(.)$ generates the output of $f_{l}(.)$ for neuron ${x}_i^{(l)}$; The neuron either remains continuously active if $f_{l,i}(\vx^{(l-1)})=\mW_{i,:}^{(l)} \vx^{(l-1)} + b_i^{(l)} > 0$, or becomes permanently inactive when $f_{l,i}(\vx^{(l-1)})=0$ or $\mW_{i,:}^{(l)}\vx^{(l-1)} + b_i^{(l)} \leq 0$. It reflects that the ReLU function is a combination of linear segments, creating a piece-wise linear function. If we constrain the neuron to remain within one of these linear segments, the ReLU function behaves linearly, making it possible to handle the proposed optimization. Therefore, the neurons in the target layer $l$ must remain in the same state (or in the same linear segment). It is formulated as follows for $\tilde{x}_i^{(l)}$:

\begin{equation}
\left\{ 
  \begin{array}{ l l }
    \tilde{\mW}_{i,:}^{(l)} \vx^{(l-1)} + \tilde{b}_i^{(l)} \geq 0 & \hspace{0.01em} \textrm{if } f_{l,i}(\vx^{(l-1)})>0,\\
    \tilde{\mW}_{i,:}^{(l)} \vx^{(l-1)} + \tilde{b}_i^{(l)} \leq 0 & \hspace{0.01em}  \textrm{otherwise.}
  \end{array}
\right.
\label{tilderelu}
\end{equation}

The above introduces a new constraint for each neuron $\tilde{x}_i^{(l)}$, where if the neuron has been originally active ($f_{l,i}(\vx^{(l-1)})=\mW_{i,:}^{(l)} \vx^{(l-1)} + b_i^{(l)} > 0$), we constrain the new weights and bias to also remain active ($\tilde{\mW}_{i,:}^{(l)} \vx^{(l-1)} + \tilde{b}_i^{(l)} \ge 0$); otherwise (if the neuron has originally been inactive, i.e., $f_{l,i}(\vx^{(l-1)})=0$ or $\mW_{i,:}^{(l)}\vx^{(l-1)} + b_i^{(l)} \leq 0$), we constrain the new weights and bias to also remain inactive ($\tilde{\mW}_{i,:}^{(l)} \vx^{(l-1)} + \tilde{b}_i^{(l)} \leq 0$).

In this framework, the values of the neurons in layers $k<l$ remain unchanged during the optimization process, but the output of layer $l$ may change. Changing neurons' values of layer $l$ may cause an alteration in the values of any neuron of the following layers $m>l$, which could potentially influence the classification result. Ensuring the consistency of the assigned class for each sample involves capturing the entire \gls{DNN}. To do so, we leverage on the piece-wise linearity of the ReLU function once more. Suppose $f_{m,j}(.)$ produces the output of $f_{m}(.)$ for neuron $\tilde{x}_j^{(m)}$. We formulate the updated value of neuron $\tilde{x}_j^{(m)}$ in layer $m>l$ as follows: 

\begin{equation}
\left\{ 
  \begin{array}{ l l }
    \mW_{j,:}^{(m)} \tilde{\vx}^{(m-1)} + b_j^{(m)} \ge 0 & \hspace{0.01em} \textrm{if } f_{m,j}(\tilde{\vx}^{(m-1)})>0,\\
    \mW_{j,:}^{(m)} \tilde{\vx}^{(m-1)} + b_j^{(m)} \leq 0   & \hspace{0.01em} \textrm{otherwise.}
  \end{array}
\right.
\label{relu}
\end{equation}

\subsection{Reformulated Layer-Wise Optimization Problem}

After relaxing the objective function and the constraints, we state our proposed layer-wise optimization problem as a linear program:

\begin{align}
&\min_{\tilde{\mW}^{(l)},\tilde{\vb}^{(l)}} {\lVert \tilde{\mW}^{(l)} \rVert}_{1,1} + {\lVert \tilde{\vb}^{(l)} \rVert}_{1}, \nonumber\\
& \hspace{0.6em}  \textrm{subject to:} \nonumber \\
&\hspace{0.8em} \vx^{(k)} = f_k(\vx^{(k-1)}), \; \forall k \in \{1, \dots, N\} \nonumber\\
& \hspace{0.0em}  \left\{ \!\!\!
  \begin{array}{ l l }
    \tilde{x}_i^{(l)} = \tilde{\mW}_{i,:}^{(l)} \vx^{(l-1)} + \tilde{b}_i^{(l)} \! \ge \! 0 & \!\!\!\! \hspace{0.2cm} \textrm{if } f_{l,i}(\vx^{(l-1)}) \! > \! 0,\nonumber\\
    \tilde{x}_i^{(l)} = 0, \tilde{\mW}_{i,:}^{(l)} \vx^{(l-1)} + \tilde{b}_i^{(l)} \! \leq \! 0   & \!\!\!\! \hspace{0.2cm} \textrm{otherwise}
  \end{array}
\right.\nonumber\\
& \hspace{0.6em}  \forall i \in \{1, \dots, n_l\}, \nonumber\\
& \hspace{0.0em}  \left\{ \!\!\! 
  \begin{array}{ l l }
    \tilde{x}_j^{(m)} \!\!= \!\! \mW_{j,:}^{(m)} \tilde{\vx}^{(m-1)} + b_j^{(m)} \!\! \ge \! 0 & \! \!\!\!\!\textrm{if }  f_{m,j}(\tilde{\vx}^{(m-1)}) \! > \! 0,\nonumber\\
    \tilde{x}_j^{(m)} \!\! = \! 0, \mW_{j,:}^{(m)} \tilde{\vx}^{(m-1)} + b_j^{(m)} \!\! \leq \! 0    &  \!\!\!\!\!\textrm{otherwise}
  \end{array}
\right.\nonumber\\
& \hspace{0.6em}  \forall j \in \{1, \dots, n_m\}, \forall m \in \{l+1, \dots, N\},\label{opt2}\\
&\hspace{0.6em}  {\lVert\tilde{\vx}^{(l)} - \vx^{(l)}\rVert}_{\infty} \leq \mathbf{{\epsilon}}, \nonumber\\
& \hspace{0.6em}  \argmax \vx^{(N)} = \argmax \tilde{\vx}^{(N)} = y,  \nonumber\\
& \hspace{0.6em}  \forall \vx^{(0)}\in \bm{D}_{val}. \nonumber
\end{align}

\subsection{End-to-End Optimization Problem}

The optimization problem in \Eqref{opt2} only targets a single layer $l$.
Here, we propose an end-to-end process to generate all layers of \glspl{VNN} iteratively. We initiate the process with an already-trained model along with a validation set. The process proceeds layer-by-layer to obtain sparse weights and biases for the \gls{DNN}. To find a verification-friendly model within our framework, we start from the first hidden layer, $l = 1$, and run the optimization problem in~\Eqref{opt2} to find the sparse matrix of weights and vector of biases such that the outcome of the classification on the validation set remains unchanged. Then, the weights and biases of the first hidden layer, $\tilde{\mW}^{(1)}$ and $\tilde{\vb}^{(1)}$, remain unaltered, while the optimization is performed on the second hidden layer to detect the sparse $\tilde{\mW}^{(2)}$ and $\tilde{\vb}^{(2)}$. This process continues until the last layer of the \gls{DNN}. 

\section{Evaluation}
In this section, we assess the performance of our proposed framework to generate \glspl{VNN} compared to state-of-the-art techniques. Our framework is implemented using the Gurobi solver~\cite{gurobi} on a MacBook Pro with an 8-core CPU and 32 GB of RAM\footnote{The code is available on \url{https://github.com/anahitabn94/VNN}.}. 

\subsection{Datasets}

We consider three public datasets for evaluation of \glspl{VNN}, namely, the MNIST dataset~\cite{mnist} and two datasets from safety-critical medical applications. The first one is about epileptic seizure detection based on the CHB-MIT Scalp EEG database~\cite{shoeb_chb-mit_2010} and the second one concerns cardiac arrhythmia detection based on the MIT-BIH Arrhythmia database~\cite{mit-bih}. 

\textbf{MNIST image dataset~\cite{mnist}} is composed of gray-scale handwritten digits such that each digit is represented by a 28$\times$28 pixel image. Similar to \cite{DeepPoly}, we consider the first 400 images of the test set and equally split them into validation and test sets for the optimization and evaluation of \glspl{DNN}, respectively.

\textbf{CHB-MIT dataset~\cite{shoeb_chb-mit_2010}} comprises 23 individuals diagnosed with epileptic seizures. This dataset is recorded in the international 10-20 system of EEG electrode positions and nomenclature. We utilize two electrode pairs, namely F7-T7 and F8-T8, which are frequently employed in seizure detection research~\cite{8351728}. The recordings are sampled at a rate of 256 Hz. For each patient, we consider 60\%, 20\%, and 20\% of the dataset for the training, validation, and test sets, respectively. The set sizes vary among patients, and each patient's entire dataset consists of 44 to 1009 samples, with an average of 252 samples.  

\textbf{MIT-BIH dataset~\cite{mit-bih}} contains ECG signals that have been digitized with a sampling rate of 360 Hz. To be able to formulate a classification problem, we select the 14 subjects with at least two distinct types of heartbeats. Similar to the CHB-MIT dataset, each model is trained using a training set, optimized with a validation set, and evaluated using a test set. The training set includes 75\% of the entire dataset. The remaining 25\% are equally partitioned among the test set and the validation set. The size of the datasets may differ among the patients and each patient's entire set consists of 400 to 1800 samples, with an average of 957 samples.

\subsection{Deep Neural Networks}

We consider \glspl{FNN} and \glspl{CNN} to investigate the performance of our proposed framework. 

\textbf{\glspl{FNN}}. We adopt \glspl{FNN} trained by~\cite{DeepPoly} and~\cite{ugare2022proof} on MNIST to explore the performance of our framework. In this paper, the notation of $n\times m$ is used for the architecture of a \gls{FNN} comprising $n$ hidden layers, each containing $m$ neurons. We examine six \glspl{FNN} with $2\times50$, $2\times100$, $5\times100$, $5\times200$, $8\times100$, and $8\times200$ architectures generated by~\cite{DeepPoly}, where each dense layer is followed by a ReLU activation function. 
Besides, we consider three \glspl{FNN} with $6\times500$ architectures, two of which are generated using \gls{PGD} adversarial training, i.e., defended against adversarial attacks~\cite{DeepPoly}. Moreover, a set of pruned \glspl{FNN} with $7\times200$ architecture, developed by~\cite{ugare2022proof}, is employed in our study. The input and output layers of all the \glspl{FNN} have 784 and 10 neurons, respectively.

\textbf{\glspl{CNN}}. We train \glspl{CNN} on CHB-MIT and MIT-BIH datasets to study the compatibility of our framework with \glspl{CNN}. For the CHB-MIT dataset, we train 23 personalized \glspl{CNN}, one for each patient. Each model consists of three hidden layers, including two convolutional layers with 3 and 5 filters and kernel sizes of 100 and 200, respectively, each followed by a max-pooling layer, as well as a dense layer with 40 neurons. The input layer has 2048 neurons. For the MIT-BIH dataset, we train 14 individual \glspl{CNN}, each with an input layer of 320 neurons, a convolutional layer including 3 filters with a kernel size of 64, and a dense layer with 40 neurons. The output layer of all \glspl{CNN} has two output neurons to classify the input into negative (non-seizure/normal) or positive (seizure/arrhythmia) classes.

\subsection{Robustness Properties}

The robustness property is formulated in terms of the $L_{\infty}$ norm and parameterized by a constant $\delta$. The adversarial region includes all perturbed inputs such that each input neuron has a maximum distance of $\delta$ from the corresponding neuron's value of the original input. The range of perturbation is considered differently for different datasets, as it depends on the robustness of each \gls{DNN} w.r.t. its structure. The maximum studied perturbations for MNIST, CHB-MIT, and MIT-BIH are 0.1, 0.01, and 0.2, respectively.

\begin{table*}[t]
    \centering
    \setlength{\tabcolsep}{5pt}
    
    \renewcommand{\arraystretch}{1.2}
    \caption{Verified robustness (\%) and average processing time (s) of original \glspl{DNN} and \glspl{VNN} ($\epsilon = 0$) on the MNIST dataset for six \glspl{FNN} with different network structures using ERAN and SafeDeep. \glspl{VNN} are optimized models that are more verification-friendly and time-efficient than their corresponding original \glspl{DNN}.}
    {
    \resizebox{\textwidth}{!}{
    \begin{tabular}{c l c c c c c c c c c c | c c c c c c c c c c }
        
        \cline{3-22}
           & & \multicolumn{10}{c}{ERAN \cite{DeepPoly}}& \multicolumn{10}{|c}{SafeDeep \cite{safedeep}}  \\
         \cline{3-22}
           & & \multicolumn{2}{c}{$\delta = 0.01$}& \multicolumn{2}{|c}{$\delta = 0.02$}&
          \multicolumn{2}{|c}{$\delta = 0.03$}&
          \multicolumn{2}{|c}{$\delta = 0.05$}& \multicolumn{2}{|c}{$\delta = 0.07$}& \multicolumn{2}{|c}{$\delta = 0.01$}&
          \multicolumn{2}{|c}{$\delta = 0.02$}& \multicolumn{2}{|c}{$\delta = 0.03$}& \multicolumn{2}{|c}{$\delta = 0.05$}& \multicolumn{2}{|c}{$\delta = 0.07$} \\
         \cline{2-22}
          & \multicolumn{1}{|l}{Model} & \multicolumn{1}{|c}{Org} & VNN &\multicolumn{1}{|c}{Org} & VNN & \multicolumn{1}{|c}{Org} & VNN & \multicolumn{1}{|c}{Org} & VNN & \multicolumn{1}{|c}{Org} & VNN &\multicolumn{1}{|c}{Org} & VNN & \multicolumn{1}{|c}{Org} & VNN & \multicolumn{1}{|c}{Org} & VNN & \multicolumn{1}{|c}{Org} & VNN & \multicolumn{1}{|c}{Org} & VNN \\
        \hline

       \multirow{6}{*}{\begin{turn}{90} \footnotesize{Robustness (\%)}\end{turn}} & \multicolumn{1}{|l}{$2 \times 50$} &\multicolumn{1}{|c}{90.5} & \textbf{92.5} 
       &\multicolumn{1}{|c}{68.5} & \textbf{85.5}
       &\multicolumn{1}{|c}{22.5} & \textbf{75.5} & \multicolumn{1}{|c}{1} & \textbf{32} & \multicolumn{1}{|c}{0.5} & \textbf{9.5} & \multicolumn{1}{|c}{91.5} & \textbf{93} &\multicolumn{1}{|c}{78} & \textbf{87.5} &\multicolumn{1}{|c}{41} & \textbf{79.5} & \multicolumn{1}{|c}{3} & \textbf{45} & \multicolumn{1}{|c}{1} & \textbf{13}  \\
       
       & \multicolumn{1}{|l}{$2 \times 100$}  & \multicolumn{1}{|c}{85} & \textbf{92.5} &\multicolumn{1}{|c}{33.5} & \textbf{85} &\multicolumn{1}{|c}{3.5} & \textbf{70} & \multicolumn{1}{|c}{0} & \textbf{16.5} & \multicolumn{1}{|c}{0} & \textbf{2} & \multicolumn{1}{|c}{90.5} & \textbf{92.5} &\multicolumn{1}{|c}{67.5} & \textbf{86.5} &\multicolumn{1}{|c}{15.5} & \textbf{77.5} & \multicolumn{1}{|c}{0} & \textbf{33.5} & \multicolumn{1}{|c}{0} & \textbf{7}  \\
        
        & \multicolumn{1}{|l}{$5 \times 100$}    & \multicolumn{1}{|c}{75.5} & \textbf{92} &\multicolumn{1}{|c}{13.5} & \textbf{74.5} &\multicolumn{1}{|c}{0.5} & \textbf{38} & \multicolumn{1}{|c}{0} & \textbf{2} & \multicolumn{1}{|c}{0} & 0 & \multicolumn{1}{|c}{87} & \textbf{92.5} &\multicolumn{1}{|c}{45} & \textbf{86} &\multicolumn{1}{|c}{8} & \textbf{67.5} & \multicolumn{1}{|c}{0} & \textbf{14.5} & \multicolumn{1}{|c}{0} & \textbf{1.5} \\
        
       & \multicolumn{1}{|l}{$5 \times 200$}& \multicolumn{1}{|c}{27} & \textbf{84} &\multicolumn{1}{|c}{0.5} & \textbf{29.5} &\multicolumn{1}{|c}{0} & \textbf{4.5} & \multicolumn{1}{|c}{0} & 0 & \multicolumn{1}{|c}{0} & 0  & \multicolumn{1}{|c}{76} & \textbf{90} &\multicolumn{1}{|c}{7} & \textbf{73} &\multicolumn{1}{|c}{0} & \textbf{27} & \multicolumn{1}{|c}{0} & \textbf{1} & \multicolumn{1}{|c}{0} & 0  \\

       & \multicolumn{1}{|l}{$8 \times 100$}  & \multicolumn{1}{|c}{67} & \textbf{85} &\multicolumn{1}{|c}{17.5} & \textbf{60.5} &\multicolumn{1}{|c}{0.5} & \textbf{28} & \multicolumn{1}{|c}{0} & \textbf{3.5} & \multicolumn{1}{|c}{0} & 0 & \multicolumn{1}{|c}{84} & \textbf{90} &\multicolumn{1}{|c}{40} & \textbf{79}  &\multicolumn{1}{|c}{11.5} & \textbf{56} & \multicolumn{1}{|c}{0} & \textbf{16.5} & \multicolumn{1}{|c}{0} & \textbf{1.5} \\
       
       & \multicolumn{1}{|l}{$8 \times 200$}  & \multicolumn{1}{|c}{22.5} & \textbf{80} &\multicolumn{1}{|c}{0.5} & \textbf{22} &\multicolumn{1}{|c}{0} & \textbf{3} & \multicolumn{1}{|c}{0} & \textbf{0.5} & \multicolumn{1}{|c}{0} & 0 & \multicolumn{1}{|c}{73} & \textbf{90} &\multicolumn{1}{|c}{5.5} & \textbf{69}  &\multicolumn{1}{|c}{0.5} & \textbf{23} & \multicolumn{1}{|c}{0} & \textbf{1.5} & \multicolumn{1}{|c}{0} & 0 \\
        \hline
        \hline

    \multirow{6}{*}{\begin{turn}{90}Average Time (s)\end{turn}} & \multicolumn{1}{|l}{$2 \times 50$}  & \multicolumn{1}{|c}{0.2} & 0.2 & \multicolumn{1}{|c}{0.2} & 0.2  &\multicolumn{1}{|c}{0.2} & 0.2 & \multicolumn{1}{|c}{0.3} & \textbf{0.2} & \multicolumn{1}{|c}{0.3} & \textbf{0.2} & \multicolumn{1}{|c}{0.8} & \textbf{0.6} & \multicolumn{1}{|c}{1.2} & \textbf{0.7} &\multicolumn{1}{|c}{1.9} & \textbf{0.7} & \multicolumn{1}{|c}{2.2} & \textbf{1.1} & \multicolumn{1}{|c}{2.2} & \textbf{1.1} \\
       
       &\multicolumn{1}{|l}{$2 \times 100$}   & \multicolumn{1}{|c}{0.5} & \textbf{0.4} & \multicolumn{1}{|c}{0.6} & \textbf{0.5}  &\multicolumn{1}{|c}{0.7} & \textbf{0.5} & \multicolumn{1}{|c}{0.8} & \textbf{0.6} & \multicolumn{1}{|c}{0.8} & \textbf{0.7} & \multicolumn{1}{|c}{2.7} & \textbf{1.7} & \multicolumn{1}{|c}{7.6} & \textbf{1.9}  &\multicolumn{1}{|c}{7.6} & \textbf{2.6} & \multicolumn{1}{|c}{10} & \textbf{4.2} & \multicolumn{1}{|c}{10} & \textbf{4.2} \\
        
         & \multicolumn{1}{|l}{$5 \times 100$}   & \multicolumn{1}{|c}{1.7} & \textbf{1.5} & \multicolumn{1}{|c}{2.3} & \textbf{1.6}  &\multicolumn{1}{|c}{2.5} & \textbf{1.9} & \multicolumn{1}{|c}{2.5} & \textbf{2.2} & \multicolumn{1}{|c}{2.5} & \textbf{2.2} & \multicolumn{1}{|c}{16} & \textbf{9} & \multicolumn{1}{|c}{36} & \textbf{13}  &\multicolumn{1}{|c}{44} & \textbf{19} & \multicolumn{1}{|c}{48} & \textbf{29} & \multicolumn{1}{|c}{48} & \textbf{30} \\
        
       & \multicolumn{1}{|l}{$5 \times 200$} & \multicolumn{1}{|c}{8} & \textbf{6} & \multicolumn{1}{|c}{9} & \textbf{7} &\multicolumn{1}{|c}{9} & \textbf{8} & \multicolumn{1}{|c}{9} & \textbf{8} & \multicolumn{1}{|c}{9} & \textbf{8}  & \multicolumn{1}{|c}{106} & \textbf{35} & \multicolumn{1}{|c}{180} & \textbf{68} &\multicolumn{1}{|c}{180} & \textbf{88} & \multicolumn{1}{|c}{180} & \textbf{94} & \multicolumn{1}{|c}{180} & \textbf{94} \\

       & \multicolumn{1}{|l}{$8 \times 100$}  & \multicolumn{1}{|c}{3} & 3 & \multicolumn{1}{|c}{4} & \textbf{3}  &\multicolumn{1}{|c}{5} & \textbf{4} & \multicolumn{1}{|c}{5} & \textbf{4} & \multicolumn{1}{|c}{5} & 5 & \multicolumn{1}{|c}{46} & \textbf{21} & \multicolumn{1}{|c}{93} & \textbf{31} &\multicolumn{1}{|c}{111} & \textbf{51} & \multicolumn{1}{|c}{123} & \textbf{60} & \multicolumn{1}{|c}{123} & \textbf{87}  \\
       
       & \multicolumn{1}{|l}{$8 \times 200$}  & \multicolumn{1}{|c}{16} & \textbf{11} & \multicolumn{1}{|c}{19} & \textbf{15} &\multicolumn{1}{|c}{20} & \textbf{17} & \multicolumn{1}{|c}{20} & \textbf{17} & \multicolumn{1}{|c}{20} & \textbf{17} & \multicolumn{1}{|c}{240} & \textbf{94} & \multicolumn{1}{|c}{325} & \textbf{140} &\multicolumn{1}{|c}{335} & \textbf{165} & \multicolumn{1}{|c}{335} & \textbf{173} & \multicolumn{1}{|c}{335} & \textbf{173} \\
         
        \hline

    \end{tabular}}
    }
    \label{tmnist_acc}
\end{table*}

\begin{figure*}[h!]
  \begin{subfigure}{0.45\textwidth}
        \centering
\includegraphics[trim={{0cm 7.4cm 0cm 7.8cm}}, clip, width=0.95\textwidth]{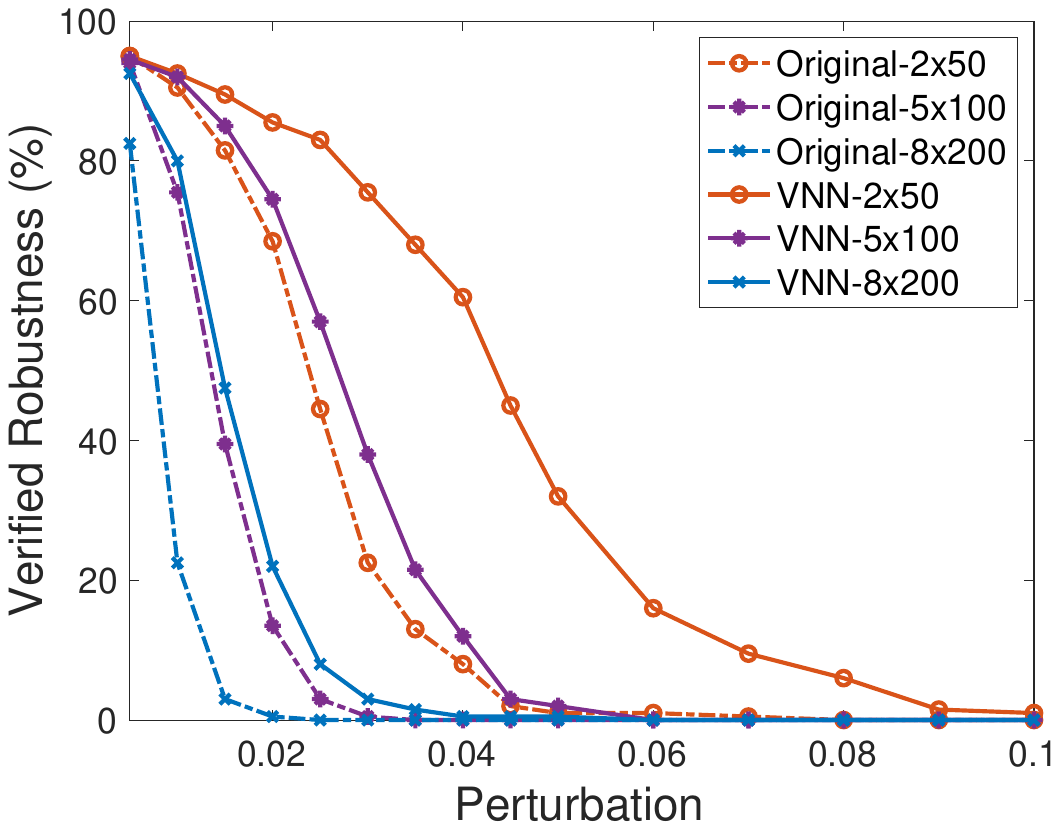}
    \caption{Verified robustness (\%) using ERAN.} \label{ERAN_acc}
  \end{subfigure} \hfill
  \begin{subfigure}{0.45\textwidth}
         \centering
         \includegraphics[trim={{0cm 7.4cm 0cm 7.8cm}}, clip, width=0.95\textwidth]{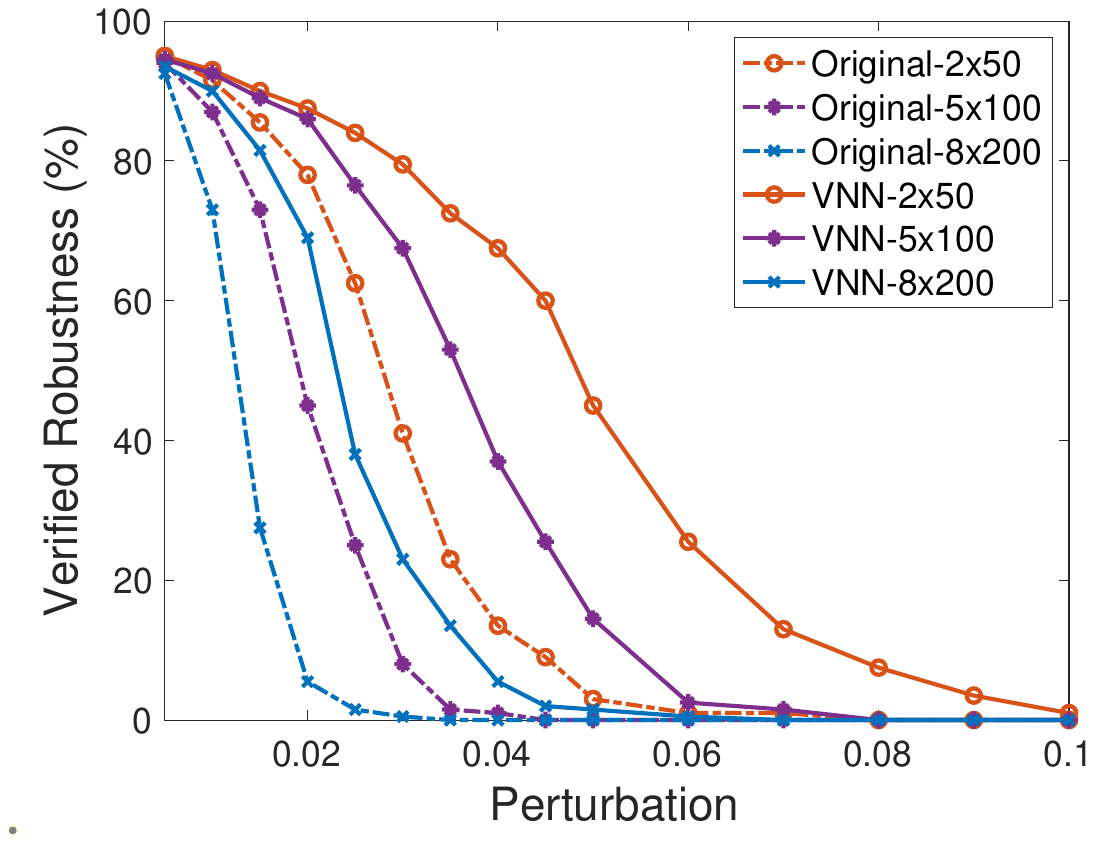}
    \caption{Verified robustness (\%) using SafeDeep.} \label{Safe_acc}
  \end{subfigure}

\caption[ caption ]
        {Verified robustness (\%) of the MNIST dataset w.r.t. different values of perturbations for original \glspl{DNN} and their corresponding \glspl{VNN} using ERAN and SafeDeep.}
         \label{MNIST}
\end{figure*}

\subsection{Verification Frameworks}
We consider ERAN~\cite{DeepPoly} and SafeDeep~\cite{safedeep} to evaluate the effectiveness of our proposed framework in this paper. ERAN and SafeDeep are over-approximation-based frameworks designed to verify the robustness of \glspl{DNN}. ERAN, or more specifically DeepPoly, works based on the principles of abstract interpretation. SafeDeep, on the other hand, works based on incremental refinement of convex approximations.

\subsection{Results and Analysis}

In this section, we evaluate our proposed framework in terms of the provability of robustness. We shall first assess the proposed \glspl{VNN} considering the MNIST dataset.

\subsubsection{MNIST.} 
We start by evaluating the \glspl{VNN} generated by our proposed framework against pre-trained original \glspl{DNN} on the MNIST dataset. \Cref{tmnist_acc} illustrates the verified robustness and average processing time of \glspl{FNN} evaluated by ERAN and SafeDeep. Here, we show five different perturbation values, $\delta$, covering a broad range of perturbations. Based on the results in~\Cref{tmnist_acc}, the number of samples whose robustness could be established is considerably larger for the \glspl{VNN} than that of the original \glspl{DNN} Moreover, as the perturbation increases, the difference in the number of verified samples becomes even more significant. Furthermore, the average processing time of \glspl{VNN} is up to three times less than their corresponding original \glspl{DNN}.

\begin{figure*}[ht]
\begin{minipage}{0.45\textwidth}
  \centering
  \includegraphics[trim={0cm 7.4cm 0cm 7.8cm}, clip, width=0.95\textwidth]{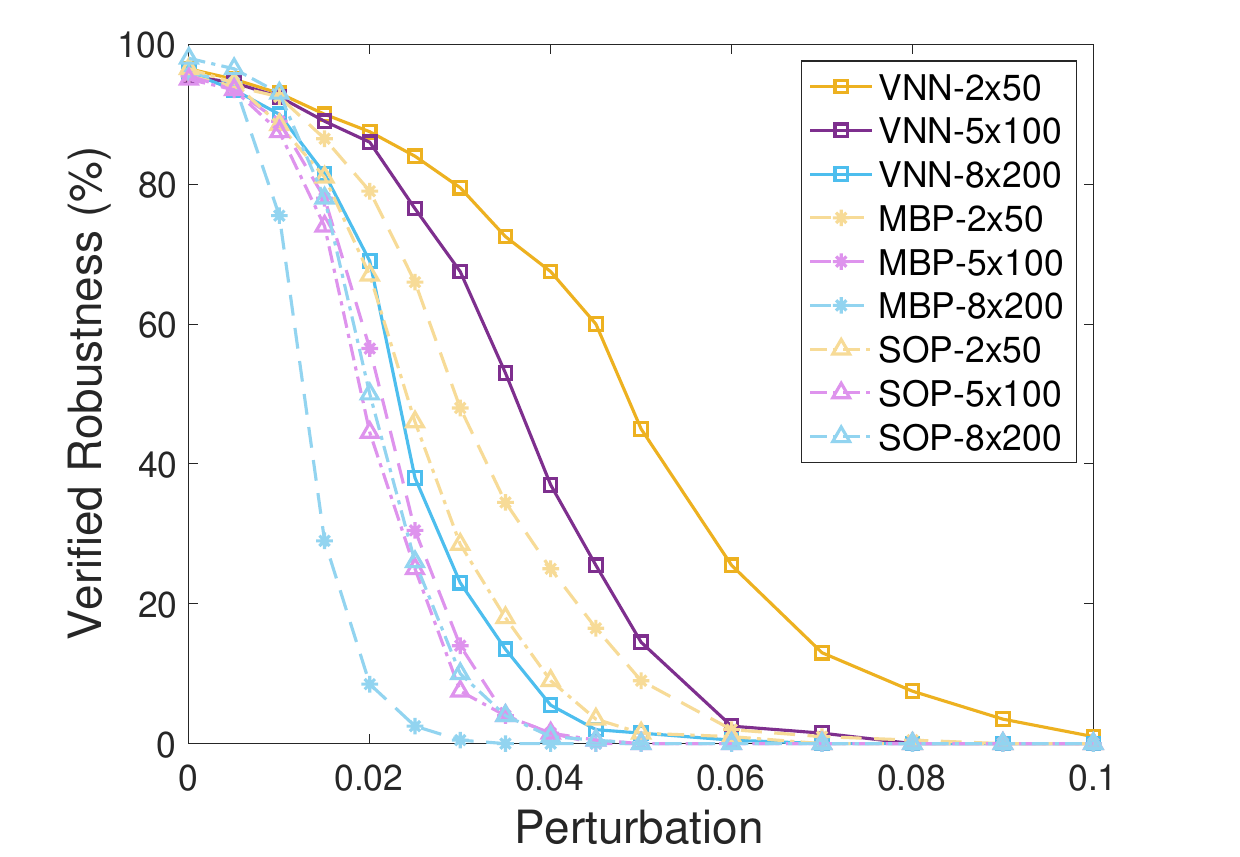}
         \caption{Comparison with state of the art: \gls{MBP} and \gls{SOP}~\cite{manngaard2018structural}.}
\label{prune}
\end{minipage}\hfill
\begin{minipage}{0.45\textwidth}
  \centering
  \includegraphics[trim={0cm 7.4cm 0cm 7.8cm}, clip, width=0.95\textwidth]{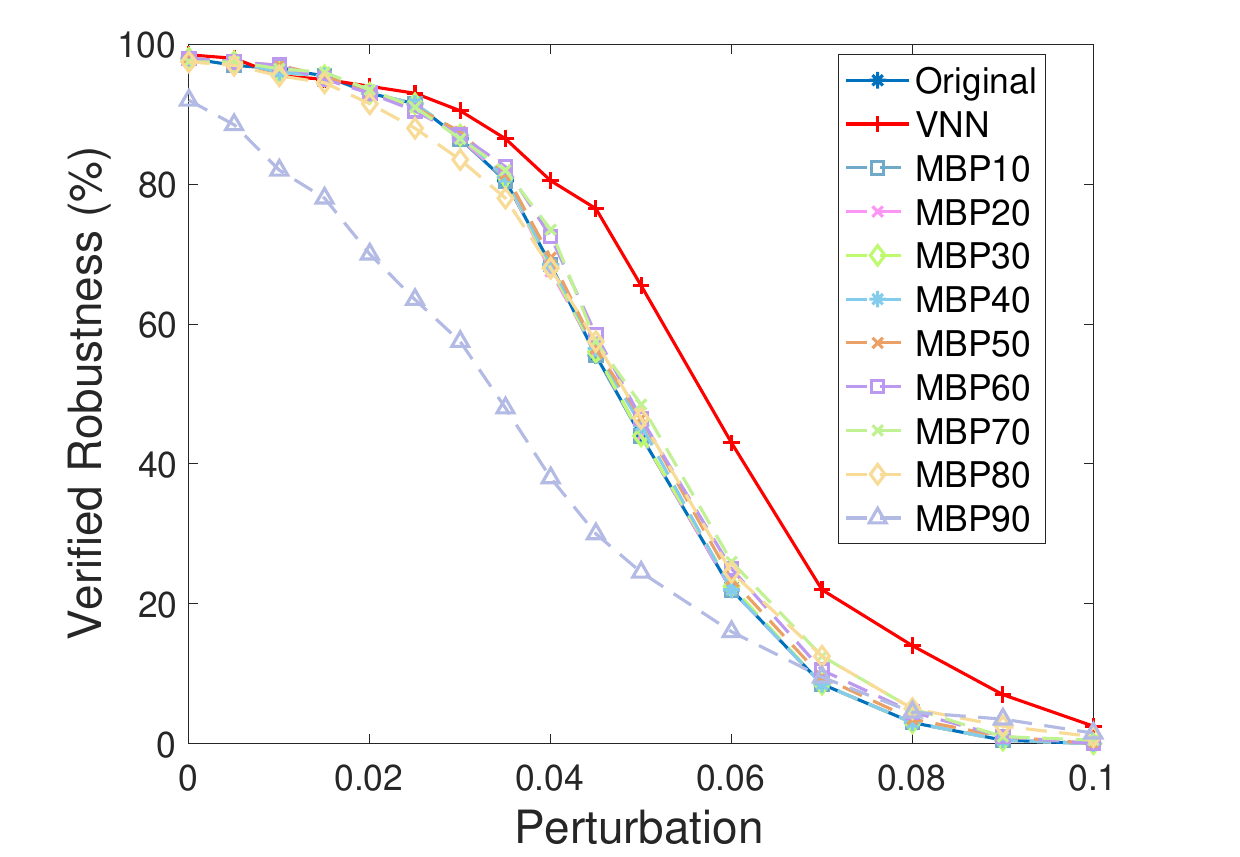}
         \caption{Comparison with state of the art: \gls{MBP}~\cite{ugare2022proof}.}
\label{prune_Ugare}
\end{minipage}
\end{figure*}

In~\Cref{MNIST} we select three \glspl{FNN} with $2 \times 50$, $5 \times 100$, and $8 \times 200$ architectures as representatives of small, medium, and large \glspl{FNN}, respectively, to study a wider range of perturbation. Based on the results in~\Cref{ERAN_acc,Safe_acc} using ERAN and SafeDeep, we observe that the accuracy of \glspl{VNN} is within the same range as their corresponding original \glspl{DNN}, which is shown in the absence of perturbation $\delta=0$. In addition, the verification tools consistently demonstrate the robustness of a greater number of samples and for larger perturbations in the case of \glspl{VNN}.

\textbf{Comparison with State-of-the-Art Pruning.} Here, we compare our framework with two state-of-the-art pruning approaches in terms of compatibility with the verification techniques: (1) \acrfull{MBP} is widely used to enhance the scalability of \glspl{DNN} by forcing small values of weights and biases to become zero~\cite{guidotti2020verification}, and (2) \acrfull{SOP} aims to enhance sparsity of \glspl{DNN} during training while preserving accuracy~\cite{manngaard2018structural}.

\Cref{prune} shows the verified robustness for three \glspl{FNN} with $2 \times 50$, $5 \times 100$, and $8 \times 200$ architectures pruned by the \gls{MBP} and \gls{SOP} compared to our proposed \glspl{VNN}. Accuracy is depicted on the y-axis where $\delta=0$. The results demonstrate that the accuracy of the \gls{MBP} and \gls{SOP} models are within the same range as \glspl{VNN}. However, the \glspl{VNN} offer substantially higher verified robustness than the corresponding \gls{MBP} and \gls{SOP} models.

Next, we conduct another experiment using the publicly available pruned \glspl{DNN} proposed by~\cite{ugare2022proof} using \gls{MBP} method. The authors shared an original and $9$ pruned \glspl{FNN} with $7 \times 200$ architecture. The pruned \glspl{DNN} are made by discarding the smallest weights at each layer with pruning rates ranging from 10\% to 90\%. The findings in \Cref{prune_Ugare} indicate that, despite their impressive accuracy shown on the y-axis where $\delta=0$, these models do not demonstrate higher verification-friendliness compared to the original one.

\textbf{Investigating Robustness of \glspl{VNN}.} 
Here, we investigate the robustness of \glspl{VNN} using the exact verification tool Marabou, an SMT-based tool~\cite{Marabou}. Exact verification frameworks, which are sound and complete, suffer from poor scalability. To overcome this issue, we only consider the \gls{FNN} with $2 \times 50$ architecture and opt for an extended timeout of 600 seconds per sample. Table~\ref{marabou_table} presents the results of the \gls{FNN} and its corresponding \gls{VNN}. For larger \glspl{FNN}, over 75\% of cases exceeded the timeout.

\begin{table}[h]
    
    \setlength{\tabcolsep}{1pt}
    \renewcommand{\arraystretch}{1.8}
    \caption{Robustness of the \gls{DNN} and \gls{VNN}.} 
    \centering
    {
    \resizebox{0.7\columnwidth}{!}{
    \begin{tabular}{ >{\centering\arraybackslash}m{0.8cm}  >{\centering\arraybackslash}m{1.6cm}  >{\centering\arraybackslash}m{1.6cm} >{\centering\arraybackslash}m{1.6cm} >{\centering\arraybackslash}m{1.6cm}  }
    
        \multicolumn{1}{c}{}& & \multicolumn{3}{c}{\gls{VNN}}\\ 
         \multicolumn{1}{c}{} & \multicolumn{1}{c}{} & \multicolumn{1}{c}{Robust} & Not Robust & Timeout \\

        \multirow{3}{*}{\begin{turn}{90}\gls{DNN}\end{turn}} & Robust & \cellcolor{black!5}41 & \cellcolor{black!5}0 & \cellcolor{yellow!45}0 \\
         &  Not Robust & \cellcolor{green!40} 269 &  \cellcolor{black!5} 275 & \cellcolor{black!5}261 \\
         & Timeout & \cellcolor{magenta!25}927 & \cellcolor{black!5}0 & \cellcolor{black!5} 27 \\
        
    \end{tabular}}
    }
    \label{marabou_table}
\end{table}

The study involves a total of 1800 states, corresponding to 200 samples in the test set, each potentially misclassified among 9 classes, as MNIST has 10 classes. The number of timeout cases in the \gls{DNN} is three times more than that in the \gls{VNN}, which implies the \gls{VNN} is more verification-friendly. On the other hand, in 269 out of the total 1800 cases, the \gls{VNN} is guaranteed to be robust, while the \gls{DNN} lacks robustness in these specific cases. Finally, no cases are identified where the \gls{DNN} exhibits robustness while the \gls{VNN} does not.

\begin{figure*}[t]
  \begin{subfigure}{.45\textwidth}
  \centering
        \includegraphics[trim={0cm 7.4cm 0cm 7.8cm}, clip, width=0.95\textwidth]{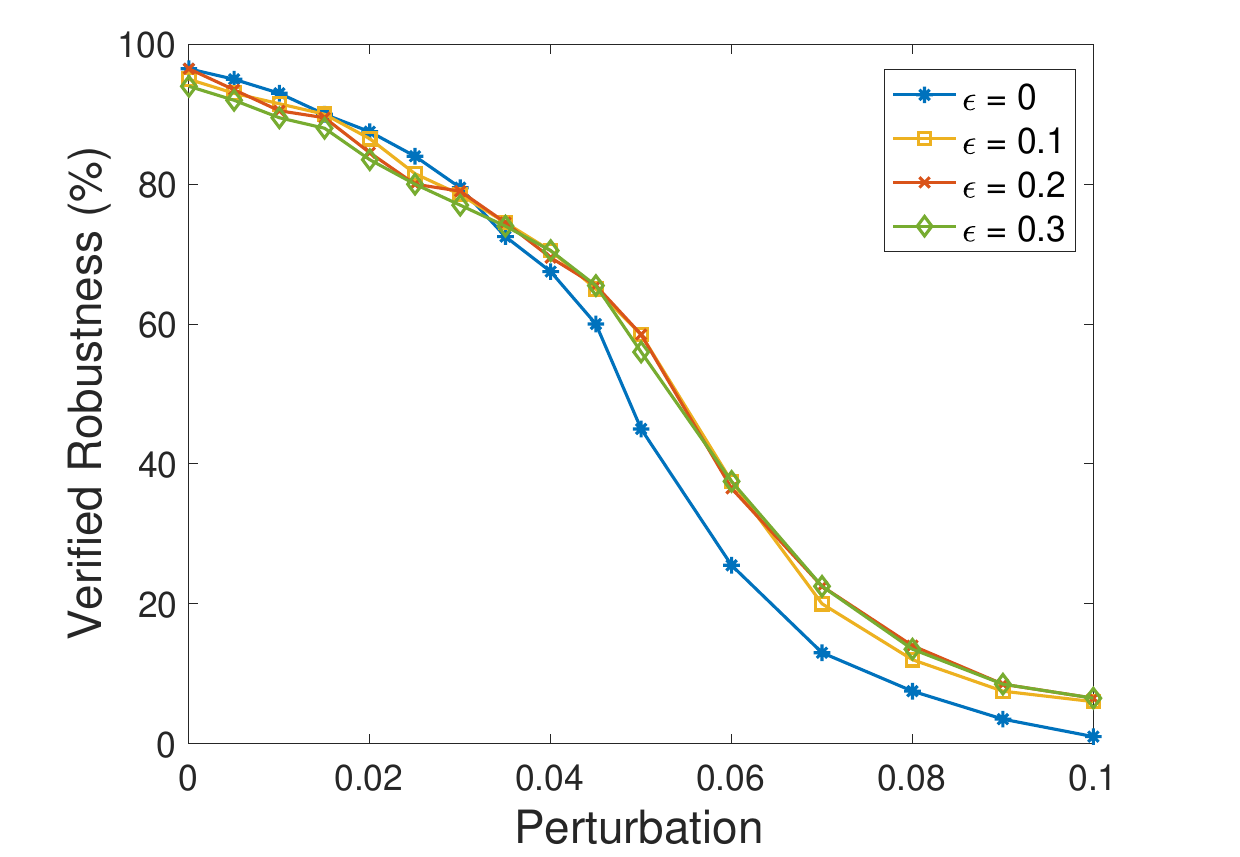}
    \caption{$2 \times 50$ \glspl{VNN}.} \label{ep1}
  \end{subfigure} \hfill
  \begin{subfigure}{.45\textwidth}
  \centering
        \includegraphics[trim={0cm 7.4cm 0cm 7.8cm}, clip, width=0.95\textwidth]{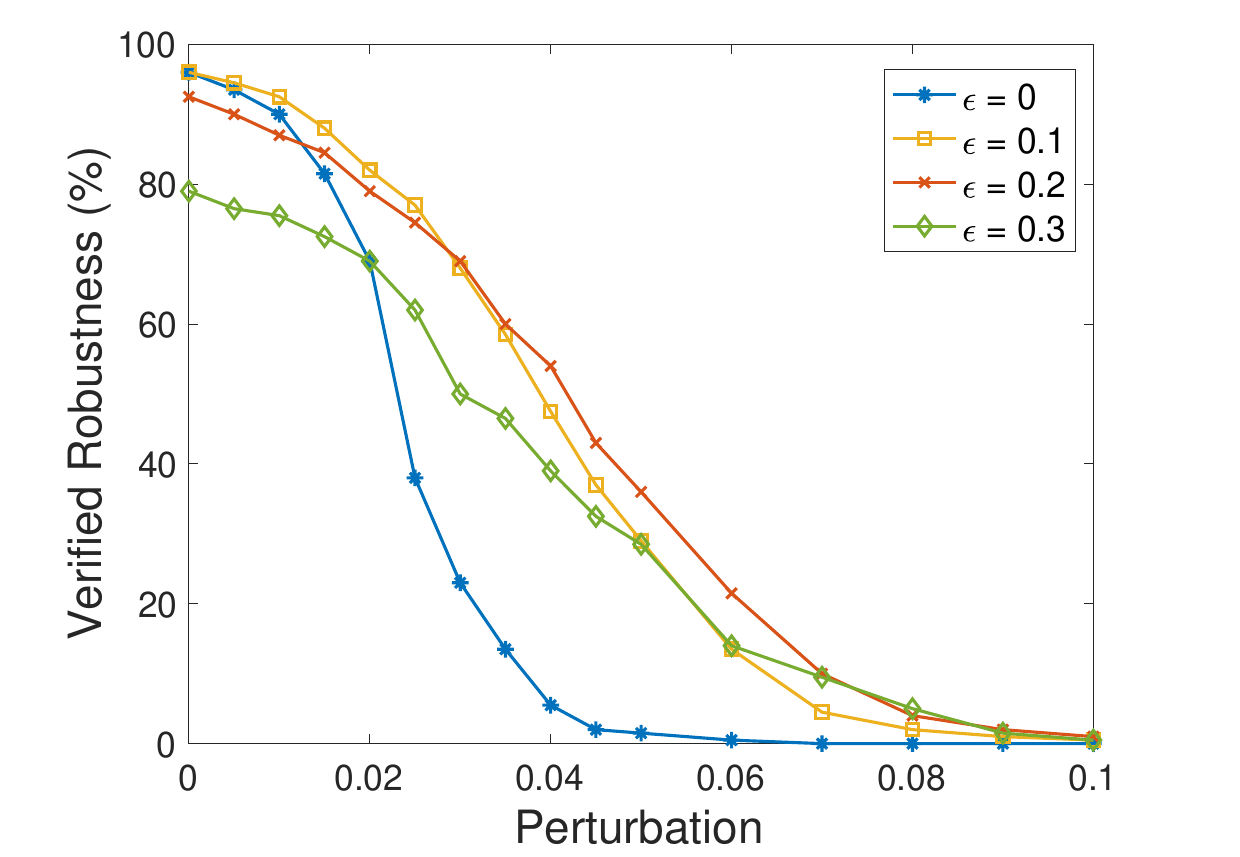}
    \caption{$8 \times 200$ \glspl{VNN}.}\label{ep3}
  \end{subfigure}
\caption[ caption ]
        {The effect of $\epsilon$ for different network structures.}
         \label{epsilons}
\end{figure*}

\textbf{Hyperparameter.} Next, we investigate how changing the hyperparameter $\epsilon$ impacts verification-friendliness on \glspl{FNN} characterized by different structures. Note that here, we employ a multiplicative perturbation. The outcomes obtained through SafeDeep are depicted in~\Cref{epsilons}, where the x-axis and y-axis represent perturbation $\delta$ and verified robustness, respectively. Note that the y-axis captures accuracy in the absence of a perturbation, which means $\delta=0$. 
The key idea is to relax the constraints, which leads to an optimization problem with a larger solution space, to obtain a sparser \gls{VNN} in the optimization problem with fewer non-zero weights/biases. For example, when $\epsilon = 0.1$, the value of each neuron $\tilde{x}_i^{(l)}$ can change in the range of $[0.9{x}_i^{(l)}, 1.1{x}_i^{(l)}]$, and the freedom of choices allows for sparser \glspl{VNN} that are easier to handle by the over-approximation-based verification techniques. However, a validation set with inadequate size, in comparison to the size of the model, may lead to a decrease in the model's prediction performance.

\Cref{ep1} displays the verification results obtained for \glspl{VNN} of a small network with two dense layers each with 50 hidden neurons. As the \gls{DNN} is small in comparison to the size of the validation set, which contains 200 samples, the accuracy of its corresponding \glspl{VNN} remains within the same range, while the verified robustness increases up to 13\%. On the other hand,~\Cref{ep3}, which illustrates the results of a large network with $9 \times 200$ architecture, shows a slight decrease in accuracy when $\epsilon$ increases. This trade-off between accuracy and verification-friendliness rises due to the small size of the validation set in relation to the sizes of the \glspl{VNN}. 
When $\epsilon=0$ or $\epsilon=0.1$, the accuracy is above 90\% and the model is more robust against small values of perturbation, while with $\epsilon=0.3$, the accuracy slightly decreases to 80\%. Our experiments show that increasing the size of the validation set increases the prediction performance.

\begin{figure*}[ht]
\begin{minipage}{0.45\textwidth}
  \centering
  \includegraphics[trim={0cm 7.4cm 0cm 7.8cm}, clip, width=0.95\textwidth]{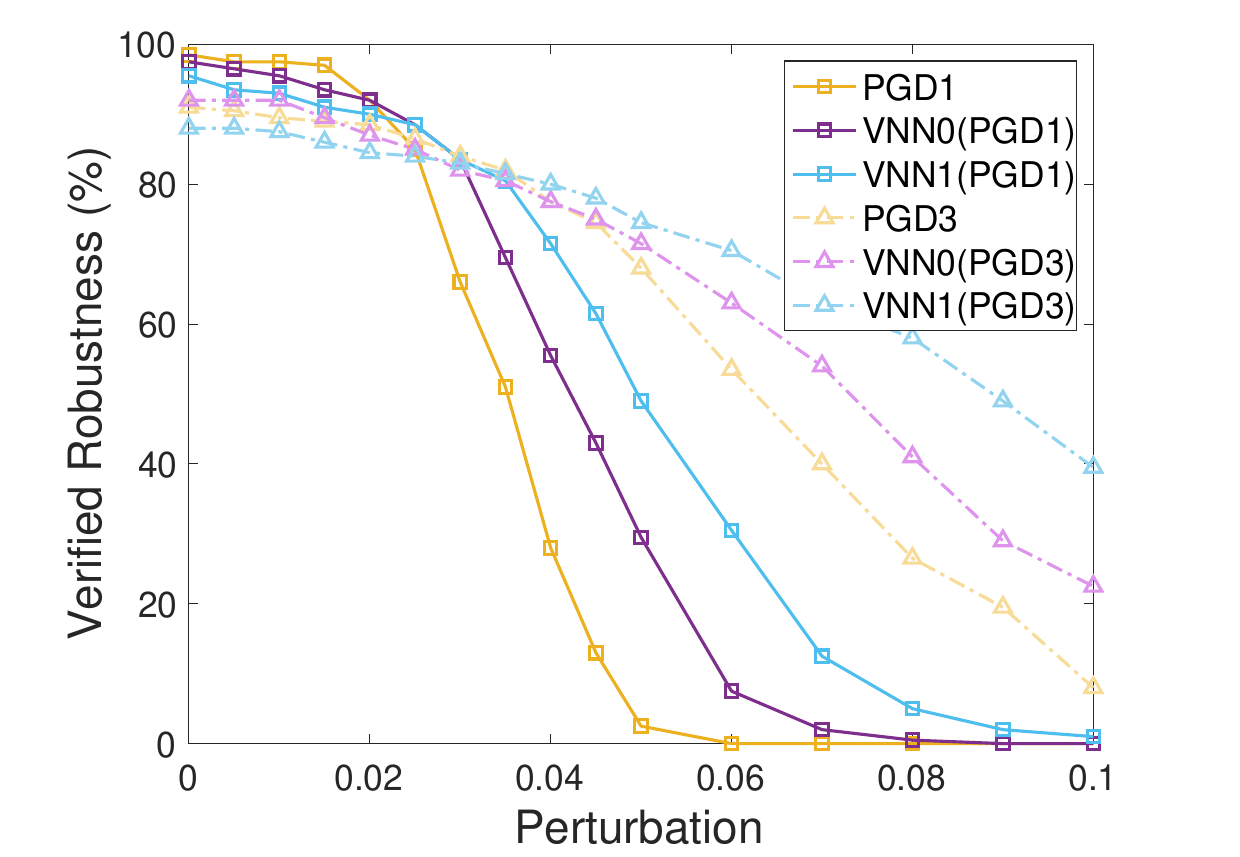}
         \caption{Comparison of \gls{PGD}-trained networks ~\cite{DeepPoly} with \gls{VNN}-enhanced networks.}
\label{pgd}
\end{minipage}\hfill
\begin{minipage}{0.45\textwidth}
  \centering
  \includegraphics[trim={0cm 7.4cm 0cm 7.8cm}, clip, width=0.95\textwidth]{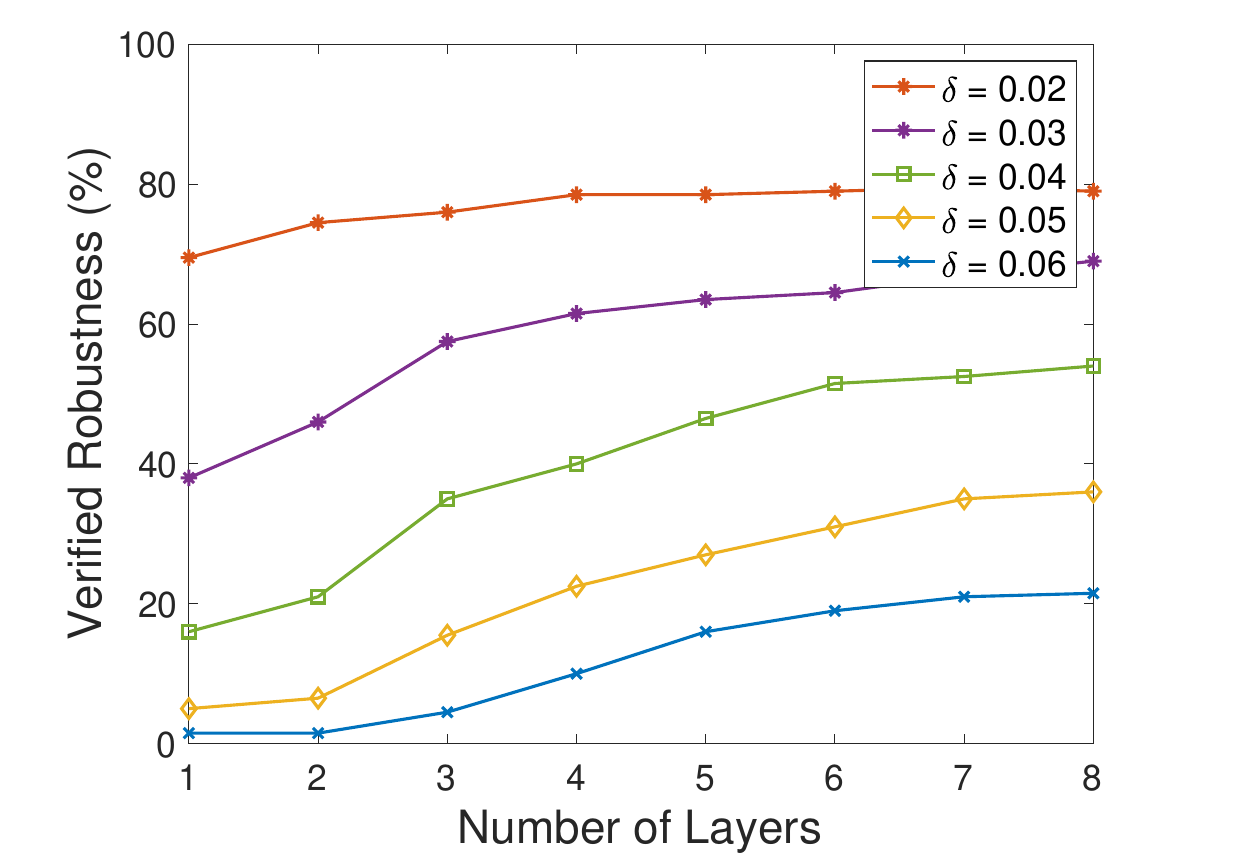}
         \caption{Increasing the number of optimized layers in an \gls{FNN} with $8 \times 200$ architecture and $\epsilon=0.2$.}  \label{layers}
\end{minipage}
\end{figure*}

\textbf{\gls{VNN} for Adversarially-Trained Models.} 
Our proposed framework can be applied in combination with state-of-the-art adversarial training techniques, which defend against adversarial attacks, such as \gls{PGD}, and still further improve verification-friendliness.

\Cref{pgd} illustrates the verified robustness for six \glspl{FNN}, all with $6 \times 500$ architectures. We consider two adversarial-trained models, trained via \gls{PGD} adversarial training with $\delta$ values of 0.1 and 0.3, as outlined in~\cite{DeepPoly}, denoted as \gls{PGD}1 and \gls{PGD}3, respectively. For each model, there are two corresponding \glspl{VNN} with $\epsilon = 0$ and $\epsilon = 0.1$. They are denoted as \gls{VNN}0(\gls{PGD}1) and \gls{VNN}1(\gls{PGD}1) for \gls{PGD}1 and \gls{VNN}0(\gls{PGD}3) and \gls{VNN}1(\gls{PGD}3) for \gls{PGD}3. The accuracy is shown on the y-axis when $\delta=0$ in \Cref{pgd}. More concretely, \Cref{pgd} shows that the accuracy values of \gls{VNN}0(\gls{PGD}1) and \gls{VNN}1(\gls{PGD}1) are comparable to that of \gls{PGD}1, while exhibiting higher verification-friendliness. The same pattern exists for the \gls{PGD}-trained model with $\delta = 0.3$, denoted as \gls{PGD}3, and its corresponding \glspl{VNN}, namely, \gls{VNN}0(\gls{PGD}3) and \gls{VNN}1(\gls{PGD}3).

\textbf{Number of Optimized Layers.} Our proposed framework optimizes the entire \gls{DNN} layer by layer. The number of optimized layers is, therefore, another parameter that affects the performance of \glspl{VNN}. ~\Cref{layers} shows the results of robustness verification for a \gls{VNN} with $8 \times 200$ architecture and $\epsilon = 0.2$ 
using SafeDeep. The x-axis shows the number of layers that are optimized using our framework, e.g., 3 means the first 3 hidden layers of the \gls{VNN} are optimized. Each curve shows the robustness verification results w.r.t. a specific value of perturbation $\delta$ to be able to compare the effect of increasing the number of optimized layers of \glspl{VNN}. \Cref{layers} demonstrates that as the number of optimized layers increases, shown on the x-axis, the proportion of verified cases increases. This phenomenon arises as increasing the number of optimized layers leads to decreasing the number of non-zero neurons and over-approximation of each neuron; thus \glspl{VNN} become more verification-friendly.

\textbf{Time Complexity.} Our proposed framework to generate \glspl{VNN} has a polynomial time complexity, since for each layer one linear program is solved. The end-to-end process of generating \glspl{VNN} also has linear time complexity with the number of layers. Experimentally, the processing time of \glspl{FNN} with $2 \times 50$,  $2 \times 100$, $5 \times 100$, $5 \times 200$, $8 \times 100$, and $8 \times 200$ architecture is 40, 101, 147, 463, 195, and 618 seconds, respectively, when $\epsilon = 0$. However, increasing the value of $\epsilon$ leads to a decrease in processing time.

Next, we evaluate our proposed framework in the context of two safety-critical medical applications, namely, epileptic seizure detection and cardiac arrhythmia detection, to demonstrate its relevance.

\subsubsection{CHB-MIT.} 
\begin{figure*}
  \begin{subfigure}[b]{.24\textwidth}
        \includegraphics[trim={0cm 0cm 0cm 0cm}, clip, width=\textwidth]{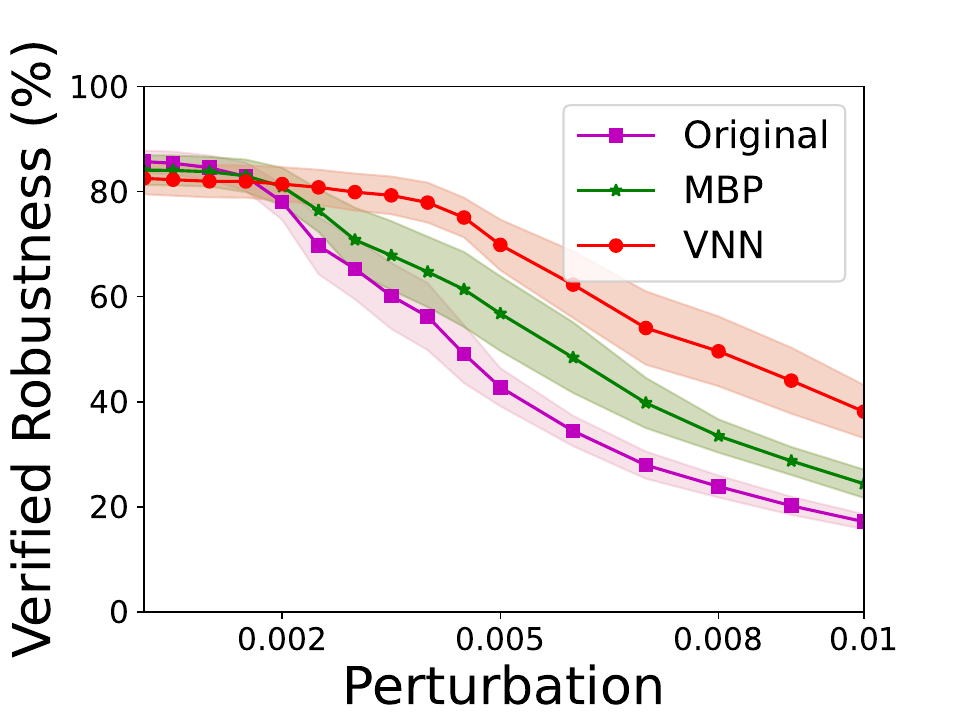}
    \caption{CHB-MIT, ERAN} \label{chberan}
  \end{subfigure}
  \begin{subfigure}[b]{.24\textwidth}
        \includegraphics[trim={0cm 0cm 0cm 0cm}, clip, width=\textwidth]{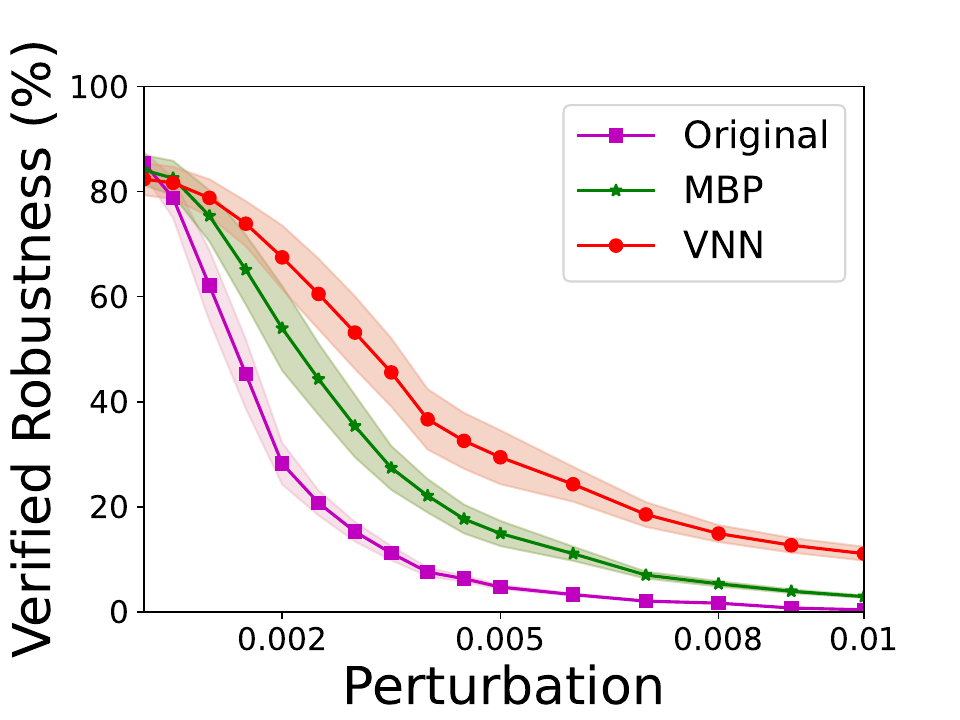}
    \caption{CHB-MIT, SafeDeep}\label{chbsd}
  \end{subfigure}
  \begin{subfigure}[b]{.24\textwidth}
        \includegraphics[trim={0cm 0cm 0cm 0cm}, clip, width=\textwidth]{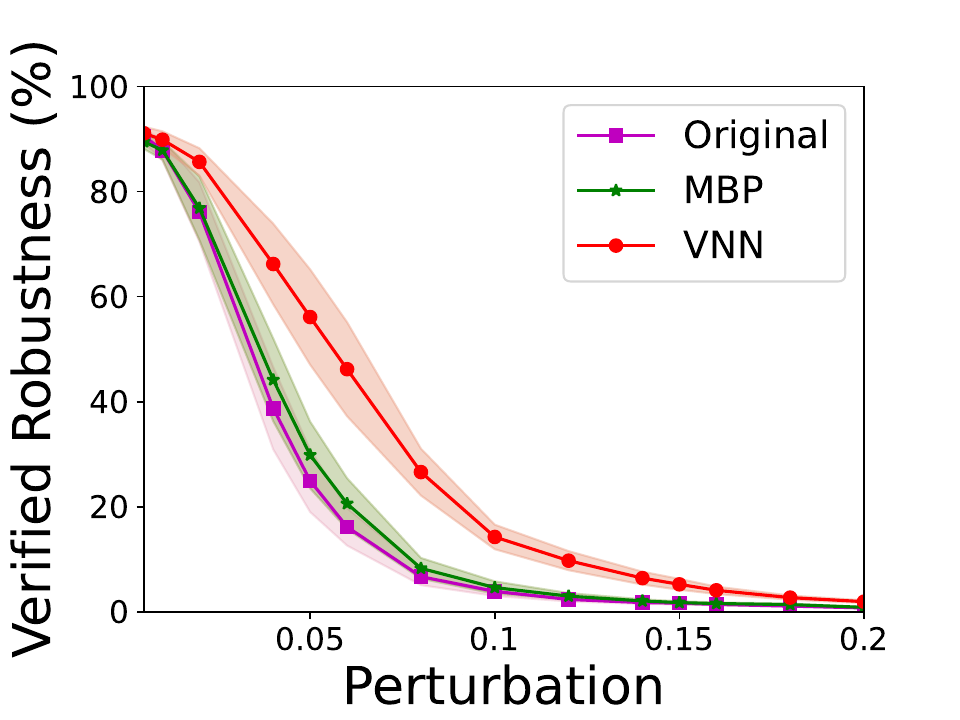}
    \caption{MIT-BIH, ERAN}\label{biheran}
  \end{subfigure}
  \begin{subfigure}[b]{.24\textwidth}
        \includegraphics[trim={0cm 0cm 0cm 0cm}, clip, width=\textwidth]{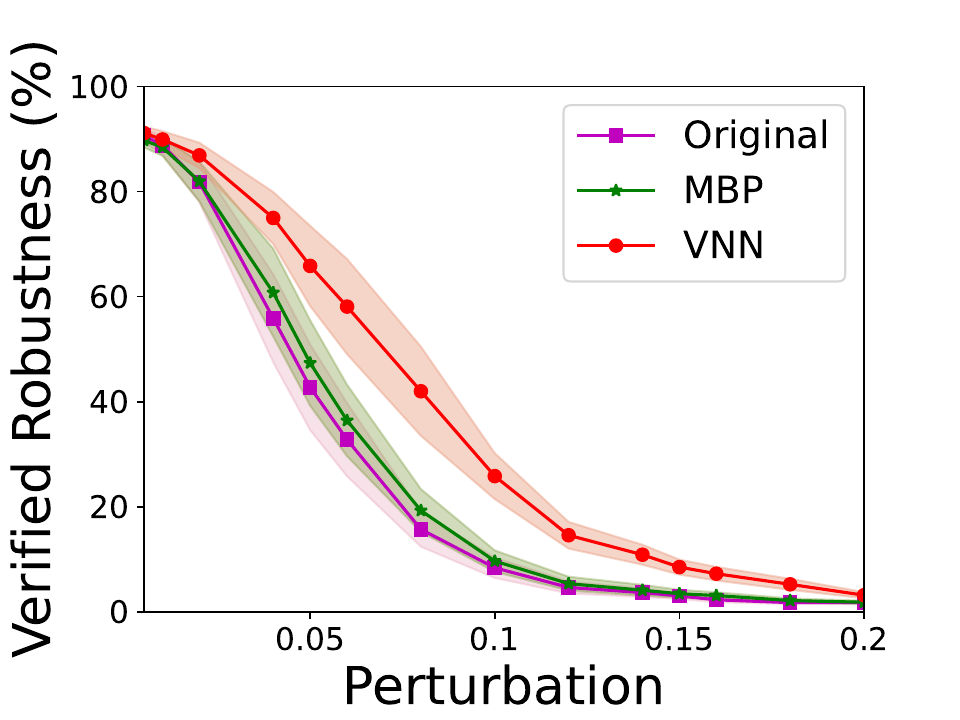}
    \caption{MIT-BIH, SafeDeep}\label{bihsd}
  \end{subfigure}
\caption[ caption ]
        {Mean and variance of verified robustness (\%) for the CHB-MIT and MIT-BIH datasets. \gls{MBP} models are generated by a threshold to force small weights to become zero, and \glspl{VNN} are generated using our proposed framework.}
         \label{CNN}
\end{figure*}

Here, we analyze our proposed framework based on the CHB-MIT dataset. \Cref{chberan,chbsd} show the verified robustness curves using ERAN and SafeDeep, where perturbation $\delta$ is in the range of $[0.0001, 0.01]$. The y-axis depicts accuracy under conditions of zero perturbation $\delta=0$. The curves demonstrate the average ratio of samples for which robustness is verified and the shaded areas represent the variance w.r.t. the patients. The verified robustness of 23 individualized \glspl{CNN} is evaluated on the patients in the CHB-MIT dataset to investigate their behavior for different perturbations. The accuracy ($\mu \pm {\sigma}^2$) of the original \glspl{CNN} and \glspl{VNN} is $85.7\% \pm 3.8\%$ and $82.5\% \pm 4.2\%$, respectively, while the \glspl{VNN} are substantially more verification-friendly. Our framework generates \glspl{VNN} with $\epsilon=0$ that have up to 9 and 24 times more verified robustness using ERAN and SafeDeep, respectively.

Moreover, we look into \gls{MBP} models, where the 10\% smallest weights and biases of the original \glspl{DNN} are set to zero ~\cite{guidotti2020verification}. By setting this threshold, 40\% to 50\% of all weights and biases became zero for each \gls{CNN}, while higher rates significantly decrease the accuracy of the \gls{MBP} models.  
Although \gls{MBP} improves verification exploring by established over-approximation-based methods, it is not as effective as our proposed optimization framework. Indeed, using a fixed threshold to disregard weights not only does not minimize the number of non-zero elements but also ignores the model's accuracy. Our proposed framework, on the other hand, actively looks for weights and biases that maintain accuracy while forcing as many as possible to become zero. For the CHB-MIT dataset, the accuracy of the \gls{MBP} models is $84.1\% \pm 4.1\%$, which is only slightly higher than \glspl{VNN}, but our framework is up to one order of magnitude more verification-friendly. This slight difference in accuracy is because \glspl{VNN} are approximately $40\%$ more sparse. However, the accuracy of the \gls{MBP} models drops to $68.4\% \pm 4.3\%$, if we enforce the \gls{MBP} models to have similar sparsity as our \glspl{VNN}.

\subsubsection{MIT-BIH.} Next, we explore the performance of our framework considering the MIT-BIH dataset using ERAN and SafeDeep. As shown in \Cref{biheran,bihsd}, \glspl{VNN} generated by our framework have a higher rate of verified robustness. Moreover, \glspl{VNN} demonstrate superior performance when subjected to higher levels of perturbation. \Cref{biheran,bihsd} show the verified robustness of original \glspl{DNN}, \gls{MBP} models, and \glspl{VNN}, with the accuracy of $91.5\% \pm 3.1\%$, $90.7\% \pm 3.4\%$, and $92.0\% \pm 3.0\%$, respectively. Note that \gls{MBP} models are created using the same conditions as CHB-MIT models, wherein a threshold is applied to set the lowest 10\% of weights and biases to zero. The experiments show that our proposed framework generates \glspl{VNN} with $\epsilon=0$ that have up to 34 and 30 times more verified robustness compared to the original \glspl{CNN} using ERAN and SafeDeep, respectively. Furthermore, the performance of \glspl{VNN} is significantly better than \gls{MBP} models.

\section{Related Work}
\glspl{DNN} are known for their vulnerability to adversarial examples~\cite{szegedy2013intriguing}. Adversarial examples are obtained by a slight modification of inputs, essentially leading to inputs that are misclassified by the \gls{DNN} despite their extreme similarity to the original correctly classified inputs. Below, we review several studies aiming at improving the scalability of robustness verification of \glspl{DNN} against adversarial examples.

One prominent research direction in this domain aims at enhancing the scalability of verification techniques for establishing robustness properties for \glspl{DNN}. Here, the state-of-the-art techniques for verification robustness of \glspl{DNN} can be categorized as either precise~\cite{MILP,Cav17,Reluplex} or based on over-approximation~\cite{DeepPoly,weng2018towards,bunel2018unified,safedeep}. The exact techniques face fundamental challenges in scalability due to their exhaustive exploration of all potential behaviors, often leading to exponential complexity~\cite{Reluplex}. On the other hand, the over-approximation-based approaches prioritize scalability over precision, but are still limited by the inherent trade-off between scalability and precision.

Other studies have sought to establish links between robustness and the architecture of \glspl{DNN}.~\citet{lin2019defensive} demonstrated that binarization enhances robustness by keeping the noise magnitude small. Another study revealed that quantized \glspl{DNN} exhibit greater scalability, but their accuracy is compromised as they disregard floating-point values~\cite{henzinger2021scalable}. In~\cite{sietsma1988neural}, it was also shown that robustness against adversarial attacks improves if redundant neurons are eliminated.~\citet{guidotti2020verification} utilized pruning to reduce the size of \glspl{DNN} by eliminating non-crucial portions that do not significantly impact their performance, but without any hard performance/robustness guarantees.~\citet{tao2023architecture} presented an approach for architecture-preserving, provable V-polytope repair of \glspl{DNN}. However, as opposed to the state-of-the-art studies above, our main focus in this paper is not on targeting specific properties, e.g., robustness; rather, the main goal of \gls{VNN}s is to enhance verifiability, while guaranteeing prediction performance/robustness requirements.

\citet{manngaard2018structural} aimed at enhancing sparsity using the Lagrange multiplier and regularizing the loss function with robustness requirements. However, when a constrained optimization problem is reformulated as an unconstrained optimization problem using the Lagrange multiplier, the final solution obtained may violate the original hard constraints. ~\citet{leofante2023verification} introduced parametric ReLU to reduce over-approximation and improve verification-friendliness, however, this method does not provide any guarantees to generate more verification-friendly \glspl{DNN} nor does it provide guarantees to satisfy performance/robustness requirements. \citet{hu2024unlocking} introduces a new residual architecture tailored for training neural networks with robustness certification but, similar to the previous work, it lacks any guarantees.

\section{Conclusions}
The state-of-the-art verification tools currently face major challenges in terms of scalability. In this paper, we presented an optimization framework to generate a new class of \glspl{DNN}, referred to as \glspl{VNN}, that are guaranteed to be accommodating to formal verification techniques. \glspl{VNN} are more time-efficient and verification-friendly while maintaining on-par prediction performance with their \gls{DNN} counterparts. We formulate an optimization problem on pre-trained \glspl{DNN} to obtain \glspl{VNN} while maintaining on-par prediction performance. Our experimental evaluation based on MNIST, CHB-MIT, and MIT-BIH datasets demonstrates that the robustness of our proposed \glspl{VNN} is established substantially more often and in a more time-efficient manner than their \gls{DNN} counterparts, without any major loss in terms of prediction performance. In addition, our experiments show that \glspl{VNN} are not only more verification-friendly, but also more robust compared to their peer \glspl{DNN}.

\section*{Acknowledgements}

This work is partially supported by the Wallenberg AI, Autonomous Systems and Software Program (WASP) funded by the Knut and Alice Wallenberg Foundation and by the European Union (EU) Interreg Program.

\section*{Impact Statement}

This paper presents work whose goal is to investigate the important task of providing formal guarantees for \glspl{DNN}, which presents a significant challenge in the AI/ML domain. This is particularly important in the context of safety-critical medical applications, e.g., epileptic seizure detection and real-time cardiac arrhythmia detection, as we show in this paper. Failure to detect an epileptic seizure or a cardiac arrhythmia episode in time may have irreversible consequences and potentially lead to death. In response to this challenge, this study introduces a framework designed to produce \glspl{DNN} that align with verification techniques, thereby enhancing the trustworthiness of these networks without sacrificing the prediction performance/robustness. Finally, we would like to highlight that there are no potential ethical impacts and future societal implications/consequences to be highlighted here.

\balance
\bibliography{bibliotek}
\bibliographystyle{icml2024}

\end{document}